\documentclass[conference]{IEEEtran}
\IEEEoverridecommandlockouts

\usepackage{array}
\usepackage[caption=false,font=large,labelfont=sf,textfont=sf]{subfig}
\usepackage{textcomp}
\usepackage{stfloats}
\usepackage{verbatim}
\usepackage{graphicx}
\usepackage{cite}
\usepackage{t1enc}

\usepackage{graphicx}
\usepackage{amsmath} 
\usepackage[english]{babel}

\usepackage{tcolorbox}
\usepackage{fancyhdr}
\usepackage{}
\usepackage{tikz}
\usepackage{multicol}
\usepackage{booktabs}
\usepackage{multirow}
\usepackage{siunitx}
\usepackage{adjustbox}
\usepackage{chngpage}
\usepackage{float}
\usepackage{comment}
\usepackage{textcomp}
\usepackage{siunitx}
\usepackage{abraces}
\usepackage[ruled,vlined]{algorithm2e}

\usepackage[utf8]{inputenc} 
\usepackage[T1]{fontenc} 
\usepackage{booktabs} 
\usepackage{amsfonts} 
\usepackage{nicefrac} 
\usepackage{microtype} 
\usepackage{amsfonts,mathtools}
\usepackage{array}
\usepackage{graphicx}
\usepackage{mathrsfs}

\usepackage{lipsum}

\usepackage{etoolbox}
\usepackage{amsmath,bm}
\usepackage{float}
\usepackage{epstopdf}
\usepackage{amssymb}

\usepackage{cite}
\usepackage{amsmath,amssymb,amsfonts}
\usepackage{graphicx}
\usepackage{textcomp}
\def\BibTeX{{\rm B\kern-.05em{\sc i\kern-.025em b}\kern-.08em
    T\kern-.1667em\lower.7ex\hbox{E}\kern-.125emX}}

\begin{document}

\title{Enhancing Robustness of Graph Neural Networks through p-Laplacian}

\author{\IEEEauthorblockN{Anuj Kumar Sirohi\textsuperscript{1} , Subhanu Halder\textsuperscript{2} , Kabir Kumar\textsuperscript{2} , Sandeep Kumar\textsuperscript{1,2} }
\IEEEauthorblockA{\textit{ \textsuperscript{1}Yardi School of Artificial Intelligence, \textsuperscript{2}Department of Electrical Engineering} \\
\textit{Indian Institute of Technology Delhi, India}\\
\{aiz218324, eez212381, ee3210741, ksandeep\}@iitd.ac.in}
}



\maketitle

\begin{abstract}
With the increase of data in day-to-day life, businesses and different stakeholders need to analyze the data for better predictions. Traditionally, relational data has been a source of various insights, but with the increase in computational power and the need to understand deeper relationships between entities, the need to design new techniques has arisen. For this graph data analysis has become an extraordinary tool for understanding the data, which reveals more realistic and flexible modelling of complex relationships. Recently, Graph Neural Networks (GNNs) have shown great promise in various applications, such as social network analysis, recommendation systems, drug discovery, and more. However, many adversarial attacks can happen over the data, whether during training (poisoning attack) or during testing (evasion attack), which can adversely manipulate the desired outcome from the GNN model. Therefore, it is crucial to make the GNNs robust to such attacks. The existing robustness methods are computationally demanding and perform poorly when the intensity of attack increases. This paper presents a computationally efficient framework, namely, $p$LapGNN, based on weighted $ p$-Laplacian for making GNNs robust. Empirical evaluation on real datasets establishes the efficacy and efficiency of the proposed method.  
\end{abstract}


\section{Introduction}
\begin{figure*}[ht]
\centering
\includegraphics[width=\textwidth,keepaspectratio]{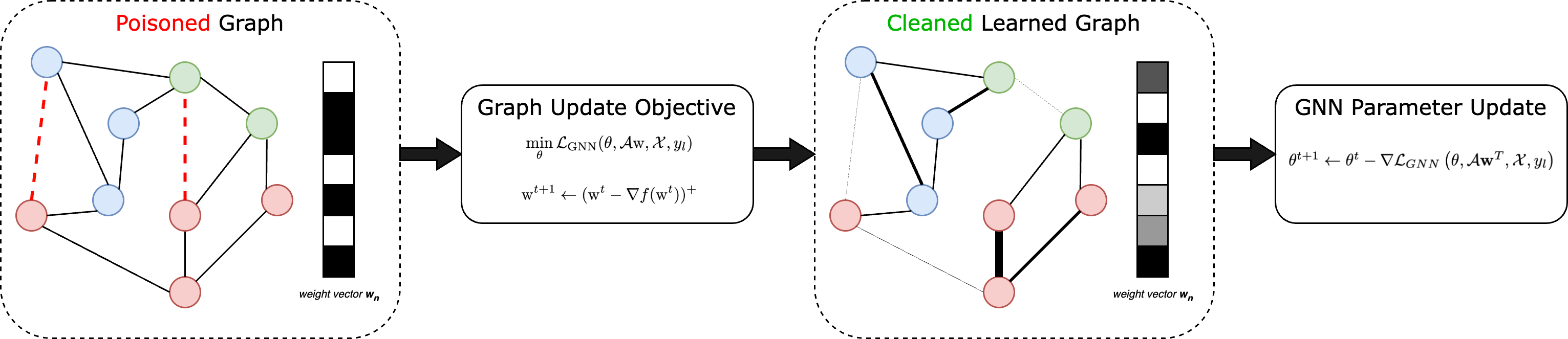}
\caption{Proposed Framework: Dashed lines depict Adversarial Edges.}
\end{figure*}

In recent times, there has been an overwhelming rise in the need to analyze the complicated relationships among entities for better analysis as graphs~\cite{wu2022graph}. Many use cases have been found in different fields, such as recommendation engines and networks involving protein and many physical structures. The edges in the graph represent essential information and the relationship between the nodes and reflect important and hidden properties of the matter that have been analyzed~\cite{wu2020comprehensive}. Hence, protecting the overall structure of the real-time graph data is very important. In recent years, GNNs have been one of the deep learning frameworks that have witnessed great success in all the above-mentioned applications and in the representation learning of graphs~\cite{wu2022graph}\cite{wu2020comprehensive} \cite{agrawal2024no}. However, the critical discussion is whether the models are capable enough to defend their properties when it is attacked. With time, different attacking mechanisms have developed to perturb a graph structure and make it lose a few of its essential properties, thus forcing it to give wrong predictions and classifications~\cite{sun2022adversarial}. The attacks make the graph heterophilic and connect nodes of distinct features. However, in the real world, the graph tends to be homophilic and has less Dirichlet energy on the graph, which forms one of the principles of retaining the original graph structure from the perturbed structure~\cite{9840814}. 

However, graphs can face various types of attacks. Such as evasion attacks occur during testing but don't affect training, which leads to incorrect test results, making them simpler for attackers. However, poisoning attacks target the training data by perturbing the adjacency matrix before training~\cite{sun2022adversarial}. These are categorized as either targeted, which poisons specific nodes for localized misclassification, or non-targeted, which impacts the entire model by poisoning all nodes. In this paper, we focus on a defence mechanism for state-of-the-art poisoning attacks, specifically \textbf{Nettack}, which is a targeted attack \cite{inproceedings} and a non-targeted attack namely, \textbf{Metattack}\cite{Zuegneradv}.

Recently, many defence mechanisms have been developed to counter adversarial attacks on GNN~\cite{pmlr-v80-dai18b}. Jaccard-GCN \cite{10.1145/3447548.3467416} preprocesses graphs by removing edges with low Jaccard similarity, addressing the homophily assumption violated in attacks. The GCN-SVD \cite{10.1145/3336191.3371789} derives a low-rank approximation of the adjacency matrix, targeting the high-frequency spectrum affected by attacks, though it risks removing original edges. ProGNN \cite{graphstructure} solves an optimization problem to recover the original adjacency matrix by leveraging the low-rank and sparse nature of graphs in general; also it introduces feature smoothing to reduce weights for dissimilar nodes. RWLGNN \cite{9840814} extends the above idea by learning the Laplacian matrix instead of the adjacency matrix, offering faster optimization through the Laplacian's properties. However, studies reveal that $p$-Laplacian has worked better in graphs, especially in heterophilic graphs and competitive performance in homophilic graphs, as it can capture local inhomogeneities more efficiently as compared to Laplacian~\cite{slepcev2019analysis} \cite{pmlr-v162-fu22e}. In this paper, we capitalize on these properties of $p$-Laplacian and propose a robustness method $p$LapGNN, in which we clean the poisoned graph using $p$-Laplacian before training GNNs. Experimental evidence on real datasets establishes the effectiveness of the proposed method as compared to baselines.

\section{Background and Problem Formulation}

In general, a graph is defined by the notation $\mathcal{G} = (\mathbb{V},\mathbb{E},\mathbb{W})$ where $\mathbb{V}$ is defined as the set containing nodes, let us consider n nodes \{$v_1$,$v_2$,....$v_n$\}. The edge set is denoted by $\mathcal{E} \subseteq \mathbb{V} \times \mathbb{V}$ and $\mathbb{W} \in \mathcal{R}^{n \times n}$ is the weighted adjacency matrix. We consider the graph is a positively weighted graph. $\forall \mathbb{W}_{ij} \geq 0$ , $\forall i \neq j$ and the graph has no self-loops. $\mathbb{W}_{ii} = 0$. There are a lot of matrix representations to capture the connections of nodes and edges in the graph \cite{9840814}  by the Laplacian or the adjacency matrix because entries of these matrix are correspond to the edges of the graph.  A matrix \(\Phi \in \mathbb{R}^{p \times p}\) is identified as a combinatorial Laplacian matrix when it belongs to the following set \cite{10.5555/3455716.3455738}:
\begin{align}\label{Lap-set}
\hspace{-1em}\mathcal{S}_{\Phi} =\big\{\Phi_{ij} =\Phi_{ji} \leq 0 \ {\rm for} \ i\neq j; \Phi_{ii}=-\sum_{j\neq i}\Phi_{ij} \big\}.
\end{align}
By construction the Laplacian matrix ($\Phi$) a positive semi-definite by structure and it has zero row and column sum, which makes the vector \textbf{1} = [1,1,...1] satisfy the property $\phi$\textbf{1} = \textbf{0} by \cite{10.5555/3455716.3455738}. Because of this properties Laplacian of the graph is most desirable for building graph based algorithms. 

Furthermore, the feature matrix is represented by the set $\mathbb{X}$ = [$x_{1}$,$x_{2}$,....,$x_{n}$] $\in \mathcal{R}^{n \times d}$ where d is the dimensions of each feature for n nodes. The notation is defined as $x_{i}$ is the feature node of $v_{i}$. \newline
In the supervised classification problem, we have fixed labels in the training data, that is used to train our model. Here we represent $\mathcal{Y}_{L}$ = \{$y_1$,$y_2$,.....,$y_l$\} as labels. We will train on a subset of nodes $\mathbb{V}_{L}$ = \{$v_1$,$v_2$,....,$v_l$\} with their corresponding labels to train the model. In the process, we try to achieve the objective of learning the function $f_\theta$, which is to  classify the unlabeled nodes
 to the correct classes function . which maps in such a way that $f_{\theta}$ : $\mathcal{V}_{l} \rightarrow \mathcal{Y}_{l}$. 
\newline
Thus the training objective function can be formulated as : 
\begin{equation} \label{GNNFormula}
\mathcal{L}_{\mathrm{GNN}} \left(\theta, \Phi, \mathbb{X}, y_{l} \right) = \sum \limits_{\textit {u}{i} \in \mathcal {V_{L}}}\textit {l}(f_{\theta }(\mathbb{X},\Phi )_{i},y_{i})
\end{equation}
Here, $l$ $\left( .,.\right)$ is the loss function such as cross-entropy and $f_{\theta }(X,\Phi )_{i}$ is the predicted lable of node $v_i$. However, in case of adverserial attacks the graph laplacian matrix $\Phi$ will be noisy or perturbed, we will denote the perturbed laplacian matrix as $\Phi_{n}$. Training the GNN model (\ref{GNNFormula}) using $\Phi_{n}$ will not yield a reliable $f_{\theta }$. In this paper our objective is to 
\textbf{first}, denoise the noisy graph laplacian $\Phi_n$ and restore a clean laplacian $\Phi^*$. The optimization formulation for obtaining the clean laplacian is: 
   \begin{gather}\label{minLnr}
     \textit{and } \quad\Phi^{*} = \arg \min _{\Phi} \mathcal{L}_{\mathrm{nr}} \left(\Phi_{n},\mathbb{X},\Phi\right)
    \end{gather}
Where, $\mathcal{L}_{\mathrm{nr}} \left(\Phi_{n},\mathbb{X},\Phi\right)$ is the noise removal objective function. 
\textbf{secondly}, train the GNN model on clean and recovered laplacian of the graph $\Phi^*$. The optimization problem we want to solve is:
    \begin{gather}\label{minlgnn}
     \min \mathcal{L}_{\mathrm{GNN}} \left(\theta, \Phi^{*}, \mathbb{X}, y_{l} \right)
     \end{gather}
in the next section we will discuss the solution of the proposed formulation \ref{minLnr} and \ref{minlgnn}.


\section{Algorithm Development}

\subsection{p-Laplacian}
$p$-Laplacian \cite{graphstructure} representation of the smoothness term can capture more intrinsic properties of the graph and derive the original graph better when it is considered a regularizer in the optimization problem of finding the original feature matrix Laplacian. The general  equation of graph laplacian can be defined by an operator, which induces the quadratic form of a function $ f: \mathcal{V} \rightarrow \mathcal{R}$ as  \cite{10.1145/1553374.1553385}. 
\begin{equation}
\left\langle f, \Delta_2 f\right\rangle=\frac{1}{2} \sum_{i, j=1}^n w_{i j}\left(f_i-f_j\right)^2
\end{equation}
The $p$-laplacian operator can be formulated as : 
\begin{equation}
\left\langle f, \Delta_p f\right\rangle=\frac{1}{2} \sum_{i, j=1}^n w_{i j}\left|f_i-f_j\right|^p .
\end{equation}

\subsection{Problem Formulation}
We define a joint optimization problem to solve (\ref{minLnr}) for  (\ref{minlgnn}) tractably and learning robust GNN parameter along with clean graph structure from input graph as \cite{9840814}:
\begin{gather}\label{minThetaPhiC}
     \min_{\theta, \Phi^{*} \in S_{\Phi}} \mathcal{L}_{\mathrm{GNN}} \left(\theta, \Phi^{*}, \mathbb{X}, y_{l} \right) +  \mathcal{L}_{nr}\left(\Phi^{*}, \Phi_{n}, \mathbb{X} \right)  
\end{gather}
where the first term is used for training GNN with clean weighted graph laplacian matrix $\Phi^{*}$  which is to be learned and the second term is the noise removal term. The noise removal objective function is defined by:
\begin{equation}
\min _{\mathbf{w} \geq 0} \mathcal{L}_{\mathrm{nr}} = \alpha\left\|\Phi^{*}-\Phi_n\right\|_F^2+\beta \sum_{i,j}\mathbf{w}_{ij}\left\|x_i - x_j\right\|_p^p
\end{equation}
Where, $\alpha\left\|\Phi^{*}-\Phi_n\right\|_F^2$ is the term used to ensure that the learned Laplacian matrix $\Phi^{*}$ does not deviate too much from the original matrix to maintain the structure and  $\alpha$ is the hyper-parameter and need to be tuned.  There are several methods available to remove edges between nodes which has dis-similar features.
Next, $\beta \sum\mathbf{w}_{ij}\left\|x_i - x_j\right\|_p^p$ ensures that we effectively remove the noise as we learn the matrix from its perturbed matrix. $p$-Laplacian offers flexibility, robustness to outliers, and the ability to promote sparsity, $\beta$ is the hyperparameter and needs to be tuned. A Dirichlet energy minimization algorithm was developed by \cite{9840814}, but the $p$-Laplacian has key advantages:
\textbf{Nonlinearity and Robustness:} With \(p > 2\), the $p$-Laplacian is more robust to outliers and better handles large feature differences between nodes by reducing dissimilar connections, unlike the Gaussian smoothing assumed by the trace term.
\textbf{Sparsity:} For \(p < 2\), it promotes sparsity, identifying key edges while ignoring weaker ones, unlike the uniform smoothing of the trace term.
\textbf{Non-Euclidean Distance Handling:} It works with non-Euclidean distances, making it suitable for graphs with irregular features, unlike the trace term.
We propose a two-stage approach where the graph is cleaned before learning GNN parameters \cite{10.1145/3336191.3371789}, \cite{DBLP:journals/corr/abs-1903-01610}, or a joint method. The two-stage approach is computationally efficient but may give suboptimal graphs, while the joint approach, though computationally demanding, is more robust at higher perturbations \cite{graphstructure}.


\textbf{Stage 1: Solving the Noise Removal Objective:} 
\begin{equation}\label{Lnrobj9} \min_{\Phi^{*} \in \mathcal{S}_{\Phi}} \mathcal{L}_{nr}(\Phi^{*}, \Phi_n, X) \end{equation} This is a Laplacian-structured constrained optimization where $\Phi \in \mathcal{S}_{\Phi}$. We simplify the matrix constraints to non-negative vector constraints using the Laplacian operator defined in \cite{10.5555/3455716.3455738}, which maps a vector $w \in \mathbb{R}^{n(n-1)/2}$ to a matrix $\mathcal{L}_w \in \mathbb{R}^{n \times n}$, satisfying Laplacian constraints ($[\mathcal{L}_w]_{ij} = [\mathcal{L}_w]_{ji}$ for $i \neq j$, and $[\mathcal{L}_w]\mathbf{1} = \mathbf{0}$).\\
\textbf{Definition 3.1:} Laplacian operator $\mathcal{L}: \mathbb{R}^{n(n-1)/2} \to \mathbb{R}^{n \times n}$, $w \mapsto \mathcal{L}_w$ is defined as: 
\[
[\mathcal{L}_w]_{ij} = 
\begin{cases} 
    -w_i + d_j & \text{if } i > j, \\
    [\mathcal{L}_w]_{ji} & \text{if } i < j, \\
    - \sum_{i \neq j} [\mathcal{L}_w]_{ij} & \text{if } i = j. 
\end{cases}
\]
where, $d_j = -j +\frac{j-1}{2}(2n-j)$\\
\textbf{Definition 3.2:} The adjoint operator $\mathcal{L}^*: Y \in \mathbb{R}^{n \times n} \to \mathcal{L}^*Y \in \mathbb{R}^{n(n-1)/2}$ is defined as: \begin{equation} [\mathcal{L}^{*} Y]k =  Y_{i,i} - Y_{i,j} - Y_{j,i} + Y_{j,j} \end{equation} 
Where, $k= i-j +\frac{j-1}{2}(2n-j) $
The operators $\mathcal{L}$ and $\mathcal{L}^*$ satisfy $\langle \mathcal{L}w, Y \rangle = \langle w, \mathcal{L}^{*}Y \rangle$.
Using $\mathcal{L}$, the Laplacian set $\mathcal{S}_\Phi$ in (1) becomes: \begin{equation}\label{Sphi} \mathcal{S}_\Phi = {\mathcal{L}_w \mid w \geq 0} \end{equation}
Replacing $\Phi^*$ with $\mathcal{L}w$ and reformulating the constraints as in \ref{Sphi}, we rewrite \ref{Lnrobj9} as:
\begin{equation}\label{reformulatedLNR} \min_{w \geq 0} \mathcal{L}_{nr} = \alpha ||\mathcal{L}w - \Phi_n||_F^2 + \beta \sum_{i,j}\mathbf{w}_{ij}\left\|x_i - x_j\right\|_p^p \end{equation}
The problem \ref{reformulatedLNR} is a non-negative constrained quadratic program: 
\begin{equation}
\min _{\mathbf{w} \geq 0} f(w) =\frac{1}{2}\left\|\mathcal{L} \mathrm{w}\right\|_F^2 - c^T.\mathrm{w} +\frac{\beta}{2\alpha}\sum\mathbf{w}_{ij}\left\|x_i - x_j\right\|_p^p
\end{equation}
where, we consider $c = 2\left(\mathcal{L}^*\Phi_n\right)^T$ \textbf{$c$} is computed beforehand before training for a dataset with any given perturbation. 
Now, due to non-negativity constraint $ w \geq 0$, the below problem does not have a closed-form solution. We use a majorization-minimization framework, where we obtain surrogate functions for objective functions such that the update rule can be obtained as in  \cite{10.5555/3455716.3455738}. Hence, we use the first-order majorization of $f(w)$ as : 
$$
g(\mathrm{w}|\mathrm{w}^{(t)}) = f(\mathrm{w}^{(t)}) + (\mathrm{w} - \mathrm{w}^{(t)})^T \nabla f(\mathrm{w}^{(t)}) + \frac{L_{1}}{2}\left\|w - w^{(t)}\right\|^2
$$
Here $L_1 = \left\|\mathcal{L}\right\|_2^2 =2n$ is Lipschitz constant. 
Now, putting the value of $f(w) =\frac{1}{2}\left\|\mathcal{L} \mathrm{w}\right\|_F^2 - c^T.\mathrm{w} +\frac{\beta}{2\alpha}\sum\mathbf{w}_{ij}\left\|x_i - x_j\right\|_p^p$ in the above equation, we have. 
$$
g(\mathrm{w}|\mathrm{w}^{(t)}) = \frac{1}{2}\mathrm{w}\mathrm{w}^T - a^T\mathrm{w} \quad\text{where,}  a = (\mathrm{w}^{(t)} - \frac{1}{L_{1}}\nabla f(\mathrm{w}^{(t)}))
$$
Using KKT conditions in the above equation, we have. 
$$
\mathrm{w}^{(t+1)} = ( \mathrm{w}^{(t)} - \frac{1}{L_{1}}\nabla f(\mathrm{w}^{(t)}) )^{+}
$$
Where, $\nabla f(w^{(t)}) = \mathcal{L}^*\left(\mathcal{L}\mathrm{w^{(t)}}\right) - c $
and we redefine the value of $c$ as 
$c = 2\mathcal{L}\left(\Phi_n\right) - \frac{\beta}{2\alpha} \sum_{i=1}^{N-1}\sum_{j=i+1}^N \left\|x_i - x_j\right\|_p^p$.This gives the updated rules for the weights. We use this algorithm to update the weight \cite{10.5555/3455716.3455738}. 
Here, $x^{t} = max(0,x)$. t is the iteration step. 
\newline
\textbf{Stage 2: GNN Parameter Learning:}
We learn the parameters of GNN using the clean graph adjacency matrix $\mathcal{A}w$
\begin{equation} \min_\theta L_{GNN}(\theta, \mathcal{A}w, X, y_l) \end{equation}

\begin{algorithm}[htbp]
\DontPrintSemicolon
\SetAlgoLined
\SetKwInOut{Input}{Input}
\SetKwInOut{Output}{Output}
\Input{$\mathcal{G} = (\Phi_n, \mathbb{X}, y_l)$, parameters: $\alpha, \beta, T, T'$}
\Output{$\mathbf{w}, \theta$}
Initialize: $\mathbf{w} \rightarrow \mathbf{w}_n, \theta \rightarrow$ Randomly\;
\BlankLine
\textbf{Stage 1:} Pre-Processing Perturbed Graph\;
\For{$t = 1$ \KwTo $T$}{
    $\mathbf{w}^{t+1} = (\mathbf{w}^t - \nabla f(\mathbf{w}^t))^+$\;
}
\BlankLine
\textbf{Stage 2:} GNN Parameter Update\;
\For{$t = 1$ \KwTo $T'$}{
    $\theta^{t+1} = (\theta^t - \nabla \mathcal{L}_{GNN}(\theta, \mathcal{A}\mathbf{w}^T, \mathbb{X}, y_l))$\;
}
\caption{$p$LapGNN }
\end{algorithm}

\section{Experiment}
In this section we evaluate the effectiveness of defense mechanism  against different kind of adversarial attacks. Before prsentating the results and observations we introduce the setup instructions:
\subsection{Experimental settings}
\subsubsection{Datasets}
We validate our $p$Lap-GNN algorithm on Cora and Citseseer, two citation network graph benchmark datasets. Table \ref{datastat} shows the statistics of both datasets.

\begin{table}[htbp!]
\caption{Description of datasets}
    \centering
    \begin{tabular}{c|cccc}
         \hline \text { Dataset } & \text { Nodes } & \text { Edges } & \text { Classes } & \text { Features } \\
\hline \text { Cora } & 2,485 & 5,069 & 7 & 1,433 \\
\text { Citeseer } & 2,110 & 3,668 & 6 & 3,703 \\
\hline

    \end{tabular}
    
    \label{datastat}
\end{table}

\subsubsection{Baseline}
We compared our method with current state-of-the-art methods ProGNN \cite{10.1145/3394486.3403049}, GNNGuard \cite{NEURIPS2020_690d8398}, RWLGNN \cite{9840814}. We employed a two-layer GCN architecture and followed the same experimental setup as \cite{graphstructure}. For each graph, we randomly selected $80\%$ of the nodes for training, $10\%$ for validation, and $10\%$ for testing. Hyper-parameters were tuned using the validation dataset. We evaluated our algorithm against two types of attacks: Targeted Attack (Nettack) and Non-Targeted Attack (Meta-self).

\subsection{Defense Performance}
To demonstrate the effectiveness of the $p$LapGNN's performance compared to the state-of-the-art defense method against different advarserial attacks we evaluate the node classification accuracy of $p$LapGNN against two type of attacks:
\begin{itemize}
    \item \textbf{Nettack:} These attacks focus on a specific subset of target nodes. We employed the state-of-the-art targeted attack, Nettack \cite{inproceedings}, for our experiments.
    \item \textbf{Metattack:} These attacks aim to degrade the overall performance of GNNs across the entire graph rather than targeting specific nodes. We used Meta-Self, a variant of the representative non-targeted attack, Metattack\cite{Zuegneradv}.    
\end{itemize}
The attack splits follow the Pro-GNN \cite{10.1145/3394486.3403049} setup. Specifically, for Nettack, nodes in the test set with a degree greater than 10 are selected as target nodes, and the number of perturbations per target node varies from 1 to 5 in steps of 1. For Metattack, perturbations range from $0\%$ to $25\%$ in steps of $5\%$. For the Random attack, random noise (addition of edges) ranges from $0\%$ to $100\%$ in steps of $20\%$.

\begin{figure}[ht]
\centering
\includegraphics[width=9cm,keepaspectratio]{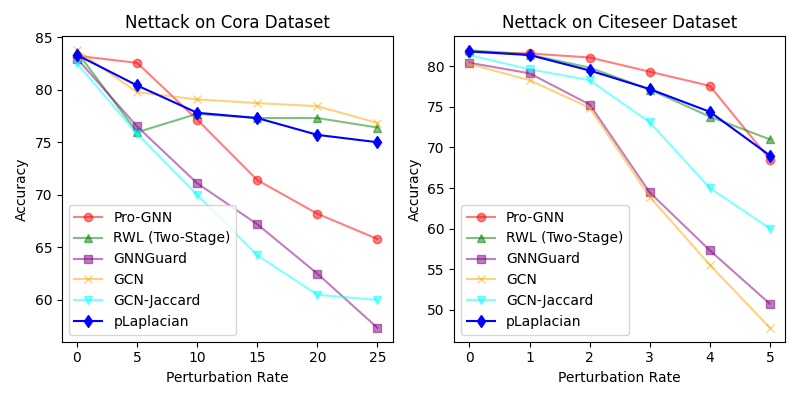}
\caption{Performance of different models under Nettack.}
\label{fig:nettack}  
\end{figure}

\begin{table}[ht]
\caption{Meta-attack analysis on Cora dataset}
    \resizebox{\columnwidth}{!}{%
    \begin{tabular}{c|c|c|c|c|c}
         \hline
        \text{Dataset} & \text{PR} & \text{Pro-GNN} & \text{Pro-GNN-two} & \text{RWL(Two-Stage)} & \text{$p$LapGNN} \\
        \hline
         & 0 & 82.98 $\pm$ 0.23 & 73.31 $\pm$ 0.71 & 83.62 $\pm$ 0.821 &  \textbf{83.75 $\pm$ 0.711} \\
        & 5 & \textbf{82.27 $\pm$ 0.45} & 73.70 $\pm$ 1.02 & 78.813 $\pm$ 0.95 & 79.76 $\pm$ 0.406 \\
        \text{Cora}& 10 & 79.03 $\pm$ 0.59 & 73.69 $\pm$ 0.81 & 79.015 $\pm$ 0.843 & \textbf{79.07 $\pm$ 0.610} \\
        & 15 & 76.40 $\pm$ 1.27 & 75.38 $\pm$ 1.10 & 78.295 $\pm$ 0.862 & \textbf{78.72 $\pm$ 0.572} \\
        & 20 & 73.32 $\pm$ 1.56 & 73.22 $\pm$ 1.08 & 78.294 $\pm$ 1.051 & \textbf{78.43 $\pm$ 0.700} \\
        & 25 & 69.72 $\pm$ 1.69 & 70.57 $\pm$ 0.61 & 76.533 $\pm$ 0.72 & \textbf{76.83 $\pm$ 0.534} \\
        \hline
    \end{tabular}
    }
    
    \label{mettatckCora}
\end{table}

\begin{table}[ht]
 \caption{Meta-attack analysis on Citeseer dataset}
    \resizebox{\columnwidth}{!}{%
    \begin{tabular}{c|c|c|c|c}
        \hline
        \text{Dataset} & \text{PR} & \text{Pro-GNN} & \text{RWL(Two-Stage)} & \text{$p$LapGNN} \\
        \hline
         & 0 & 72.28 $\pm$ 0.69 & 70.842 $\pm$ 0.454 & \textbf{73.08 $\pm$ 0.73} \\
        & 1 & 71.93 $\pm$ 0.57 & 71.103 $\pm$ 0.865 & \textbf{ 72.44 $\pm$ 0.55 }\\
        \text{Citeseer}& 2 & 70.51 $\pm$ 0.75 & 70.263 $\pm$ 1.015 & \textbf{ 71.02 $\pm$ 0.71} \\
        & 3 & 69.03 $\pm$ 1.11 & 68.453 $\pm$ 1.306 &  \textbf{69.52 $\pm$ 0.76} \\
        & 4 & 68.02 $\pm$ 2.28 & 67.927 $\pm$ 1.231 & \textbf{ 68.54 $\pm$ 0.87 }\\
        & 5 & 68.95 $\pm$ 2.78 & \textbf{70.433 $\pm$ 0.763} & 69.01 $\pm$ 0.53 \\
        \hline
    \end{tabular}
    }
   
    \label{mettatcakCiteseer}
\end{table}
The $p$-Laplacian has key advantages for node classification. We present node classification accuracy results averaged over $10$ runs for Metattack and $5$ runs for Nettack on the Cora and Citeseer datasets under various attacks and perturbation rates. The proposed $p$LapGNN consistently outperforms others across most cases. Notably, $p$LapGNN converges significantly faster, requiring only $200$ epochs in stage $1$ preprocessing, compared to $1000$ epochs for Pro-GNN (Joint) under Nettack, making it much more efficient in training time.


\subsection{Parameter Tuning and Libraries}
We have used deeprobust library for implementing different attacks and GCN architecture as well. The splits are done as per the ProGNN \cite{10.1145/3394486.3403049}. The nodes in the test set with degrees having values higher than 10 are selected for target node in Nettattack. The perturbation in Meta ranges from 0\% to 25\% with a step of 5\%. In nettatack, the attacks are varied from 1 to 5 with a step of 1. The optimizer used was SGD for updating the weights with the gradient. The learning rate for optimizing the weights was selected as $1e^{-3}$ and for the GNN model was $1e^{-2}$. We used 200 epochs for solving the optimization problem to find back the de-noised laplacian matrix, while for the GNN model to train, we used around 250 epochs. We did extensive testing for finding the value of $\beta$ and we tried for values from 0.1 to 1.5 for all the perturbations. While we also varied the value of p to find which value of p gives us a better value. Also, we fixed $\alpha$ = 1 which worked the best after testing. 

\subsection{Complexity and Run-time Analysis}
The proposed framework $p$LapGNN, has minimal computational overhead and integrates easily with GNN architectures like GCN, which has a complexity of $O(|\mathbb{V}| + |\mathbb{E}|)$. In contrast, defence methods viz. ProGNN is based on SVD decomposition, which is computationally expensive at $O(|\mathbb{V}|^3)$.


\section{Conclusion}
Graph neural networks are vulnerable to adversarial attacks, and it is important to restore the original graph even if it has been perturbed. In this work, we took a holistic approach based on $p$-Laplacian to the problem, to restore not only the original structure but also to make the model learn about the structure so that it can help give better predictions. Empirical evaluation of our proposed framework $p$LapGNN shows competitive performance while converging much faster as compared to baselines. Its performance is better or at par with the state-of-the-art method on random attacks while slightly lower ($1-2\%$) on targeted attacks. In our future work, we would like to improve the existing framework by trying a single-stage optimization method and achieving a more stable GNN performance for consistent accuracies and better performance over all types of attacks.

\bibliographystyle{IEEEtran}
\bibliography{reference.bib, ref_extra}

\end{document}